\documentclass{article}

\usepackage{myaclap,psfig,dialog2e,times,new-algorithm,subfigure}

\title{Tracking Initiative in Collaborative Dialogue Interactions}

\author{Jennifer Chu-Carroll \and Michael K. Brown \\
        Bell Laboratories \\
        Lucent Technologies \\
        600 Mountain Avenue \\
        Murray Hill, NJ 07974, U.S.A.\\
        E-mail: \{jencc,mkb\}@bell-labs.com}

\date{}

\begin{document}

\renewcommand{\floatpagefraction}{0}

\maketitle

\begin{abstract}

In this paper, we argue for the need to distinguish between {\em task} and
{\em dialogue} initiatives, and present a model for tracking shifts in both
types of initiatives in dialogue interactions.  Our model predicts the
initiative holders in the next dialogue turn based on the current
initiative holders and the effect that observed cues have on changing
them. Our evaluation across various corpora shows that the use of cues
consistently improves the accuracy in the system's prediction of task and
dialogue initiative holders by 2-4 and 8-13 percentage points,
respectively, thus illustrating the generality of our model.

\end{abstract}

\section{Introduction}

Naturally-occurring collaborative dialogues are very rarely, if ever,
one-sided. Instead, initiative of the interaction shifts among participants
in a primarily principled fashion, signaled by features such as linguistic
cues, prosodic cues and, in face-to-face interactions, eye gaze and
gestures. Thus, for a dialogue system to interact with its user in a
natural and coherent manner, it must recognize the user's cues for
initiative shifts and provide appropriate cues in its responses to user
utterances.

Previous work on mixed-initiative dialogues focused on tracking a single
thread of control among participants. We argue that this view of initiative
fails to distinguish between {\em task initiative} and {\em dialogue
initiative}, which together determine when and how an agent will address an
issue.  Although physical cues, such as gestures and eye gaze, play an
important role in coordinating initiative shifts in face-to-face
interactions, a great deal of information regarding initiative shifts can
be extracted from utterances based on linguistic and domain knowledge
alone. By taking into account such cues during dialogue interactions, the
system is better able to determine the task and dialogue initiative holders
for each turn and to tailor its response to user utterances accordingly.

In this paper, we show how distinguishing between task and dialogue
initiatives accounts for phenomena in collaborative dialogues that previous
models were unable to explain. We show that a set of cues, which can be
recognized based on linguistic and domain knowledge alone, can be utilized
by a model for tracking initiative to predict the task and dialogue
initiative holders with 99.1\% and 87.8\% accuracies, respectively, in
collaborative planning dialogues. Furthermore, application of our model to
dialogues in various other collaborative environments consistently
increases the accuracies in the prediction of task and dialogue initiative
holders by 2-4 and 8-13 percentage points, respectively, compared to a
simple prediction method without the use of cues, thus illustrating the
generality of our model.

\section{Task Initiative vs. Dialogue Initiative}

\subsection{Motivation}

Previous work on mixed-initiative dialogues focused on tracking and
allocating a single thread of control, the {\em conversational lead}, among
participants. Novick \shortcite{nov_phd} developed a computational model
that utilizes {\em meta-locutionary acts}, such as {\em repeat} and {\em
give-turn}, to capture mixed-initiative behavior in dialogues. Whittaker
and Stenton \shortcite{whi_ste_acl88} devised rules for allocating dialogue
control based on utterance types, and Walker and Whittaker
\shortcite{wal_whi_acl90} utilized these rules for an analytical study on
discourse segmentation. Kitano and Van Ess-Dykema \shortcite{kit_ved_acl91}
developed a plan-based dialogue understanding model that tracks the
conversational initiative based on the domain and discourse plans behind
the utterances. Smith and Hipp \shortcite{smi_hip_book94} developed a
dialogue system that varies its responses to user utterances based on four
dialogue modes which model different levels of initiative exhibited by
dialogue participants. However, the dialogue mode is determined at the
outset and cannot be changed during the dialogue. Guinn
\shortcite{gui_acl96} subsequently developed a system that allows change in
the level of initiative based on initiative-changing utterances and each
agent's competency in completing the current subtask.

However, we contend that merely maintaining the conversational lead is
insufficient for modeling complex behavior commonly found in
naturally-occurring collaborative dialogues
\cite{sri92,trains93,trains93a}. For instance, consider the alternative
responses in utterances (\ref{smith})-(\ref{other}), given by an advisor to
a student's question:

\bdialog{A:}{S:}
\em

\listenerlab I want to take NLP to satisfy my seminar course
requirement.

\dialine \label{nlp} Who is teaching NLP?

\speakerlabsub \label{smith} Dr.~Smith is teaching NLP.

\speakerlabsubcont \label{can't} You can't take NLP because you haven't
taken AI, which is a prerequisite for NLP.

\speakerlabsubcont \label{other} You can't take NLP because you haven't
taken AI, which is a prerequisite for NLP. You should take distributed
programming to satisfy your requirement, and sign up as a listener for NLP.

\edialog

Suppose we adopt a model that maintains a single thread of control, such as
that of \cite{whi_ste_acl88}. In utterance (\ref{smith}), A directly
responds to S's question; thus the conversational lead remains with S. On
the other hand, in (\ref{can't}) and (\ref{other}), A takes the lead by
initiating a subdialogue to correct S's invalid proposal.  However,
existing models cannot explain the difference in the two responses, namely
that in (\ref{other}), A actively participates in the planning process by
explicitly proposing domain actions, whereas in (\ref{can't}), she merely
conveys the invalidity of S's proposal. Based on this observation, we argue
that it is necessary to distinguish between {\em task initiative}, which
tracks the lead in the development of the agents' plan, and {\em dialogue
initiative}, which tracks the lead in determining the current discourse
focus \cite{chu_bro_sss97}.\footnote{Although independently conceived, this
distinction between task and dialogue initiatives is similar to the notion
of {\em choice of task} and {\em choice of speaker} in initiative in
\cite{nov_sut_sss97}, and the distinction between {\em control} and {\em
initiative} in \cite{jor_die_sss97}.} This distinction then allows us to
explain A's behavior from a response generation point of view: in
(\ref{can't}), A responds to S's proposal by merely taking over the
dialogue initiative, i.e., informing S of the invalidity of the proposal,
while in (\ref{other}), A responds by taking over both the task and
dialogue initiatives, i.e., informing S of the invalidity and suggesting a
possible remedy.

\looseness=-1000 An agent is said to have the {\em task initiative} if she
is directing how the agents' task should be accomplished, i.e., if her
utterances directly propose {\em actions} that the agents should
perform. The utterances may propose {\em domain} actions
\cite{lit_all_cs87} that directly contribute to achieving the agents' goal,
such as {\em ``Let's send engine E2 to Corning.''} On the other hand, they
may propose {\em problem-solving} actions
\cite{all_snlw91,lam_car_acl91,ram_acl91} that contribute not directly to
the agents' domain goal, but to how they would go about achieving this
goal, such as {\em ``Let's look at the first [problem] first.''} An agent
is said to have the {\em dialogue initiative} if she takes the
conversational lead in order to establish mutual beliefs, such as mutual
beliefs about a piece of domain knowledge or about the validity of a
proposal, between the agents. For instance, in responding to agent A's
proposal of sending a boxcar to Corning via Dansville, agent B may take
over the dialogue initiative (but not the task initiative) by saying ``{\em
We can't go by Dansville because we've got Engine 1 going on that track.}''
Thus, when an agent takes over the task initiative, she also takes over the
dialogue initiative, since a proposal of actions can be viewed as an
attempt to establish the mutual belief that a set of actions be adopted. On
the other hand, an agent may take over the dialogue initiative but not the
task initiative, as in (\ref{can't}) above.

\subsection{An Analysis of the TRAINS91 Dialogues}

To analyze the distribution of task/dialogue initiatives in collaborative
planning dialogues, we annotated the TRAINS91 dialogues \cite{trains93} as
follows: each dialogue turn is given two labels, {\em task initiative (TI)}
and {\em dialogue initiative (DI)}, each of which can be assigned one of
two values, {\em system} or {\em manager}, depending on which agent holds
the task/dialogue initiative during that turn.\footnote{An agent holds the
task initiative during a turn as long as {\em some} utterance during the
turn directly proposes how the agents should accomplish their goal, as in
utterance (\ref{other}).}

\begin{table}[tb]
\footnotesize
\begin{center}
\begin{tabular}{|l|c|c|}
\hline
 & {\bf TI: system} & {\bf TI: manager} \\ \hline
{\bf DI: system} & 37 (3.5\%) & 274 (26.3\%) \\ \hline
{\bf DI: manager} & 4 (0.4\%) & 727 (69.8\%) \\ \hline
\end{tabular}
\end{center}
\caption{Distribution of Task and Dialogue Initiatives}
\label{distribution}
\vspace{1em}
\hrule
\end{table}

Table~\ref{distribution} shows the distribution of task and dialogue
initiatives in the TRAINS91 dialogues. It shows that while in the majority
of turns, the task and dialogue initiatives are held by the same agent, in
approximately 1/4 of the turns, the agents' behavior can be better
accounted for by tracking the two types of initiatives separately.

To assess the reliability of our annotations, approximately 10\% of the
dialogues were annotated by two additional coders. We then used the kappa
statistic \cite{sie_cas_book88,car_cl96} to assess the level of agreement
between the three coders with respect to the task and dialogue initiative
holders. In this experiment, $K$ is 0.57 for the task initiative holder
agreement and $K$ is 0.69 for the dialogue initiative holder agreement.

Carletta suggests that content analysis researchers consider $K>$.8 as good
reliability, with .67$<K<$.8 allowing tentative conclusions to be drawn
\cite{car_cl96}. Strictly based on this metric, our results indicate that
the three coders have a reasonable level of agreement with respect to the
dialogue initiative holders, but do not have reliable agreement with
respect to the task initiative holders. However, the kappa statistic is
known to be highly problematic in measuring inter-coder reliability when
the likelihood of one category being chosen overwhelms that of the other
\cite{groetal_agp81}, which is the case for the task initiative
distribution in the TRAINS91 corpus, as shown in
Table~\ref{distribution}. Furthermore, as will be shown in
Table~\ref{domains}, Section~\ref{apps}, the task and dialogue initiative
distributions in TRAINS91 are not at all representative of collaborative
dialogues. We expect that by taking a sample of dialogues whose
task/dialogue initiative distributions are more representative of all
dialogues, we will lower the value of P(E), the probability of chance
agreement, and thus obtain a higher kappa coefficient of
agreement. However, we leave selecting and annotating such a subset of
representative dialogues for future work.

\section{A Model for Tracking Initiative}

Our analysis shows that the task and dialogue initiatives shift between the
participants during the course of a dialogue. We contend that it is
important for the agents to take into account signals for such initiative
shifts for two reasons. First, recognizing and providing signals for
initiative shifts allow the agents to better coordinate their actions, thus
leading to more coherent and cooperative dialogues. Second, by determining
whether or not it should hold the task and/or dialogue initiatives when
responding to user utterances, a dialogue system is able to tailor its
responses based on the distribution of initiatives, as illustrated by the
previous dialogue \cite{chu_bro_sss97}. This section describes our
model for tracking initiative using cues identified from the
user's utterances.

Our model maintains, for each agent, a {\em task initiative index} and
a {\em dialogue initiative index} which measure the amount of evidence
available to support the agent holding the task and dialogue
initiatives, respectively. After each turn, new initiative indices are
calculated based on the current indices and the effects of the cues
observed during the turn. These cues may be explicit requests by the
speaker to give up his initiative, or implicit cues such as ambiguous
proposals. The new initiative indices then determine the initiative
holders for the next turn.

We adopt the Dempster-Shafer theory of evidence
\cite{sha_book76,gor_sho_mycin84} as our underlying model for inferring the
accumulated effect of multiple cues on determining the initiative indices.
The Dempster-Shafer theory is a mathematical theory for reasoning under
uncertainty which operates over a set of possible outcomes, $\Theta$.
Associated with each piece of evidence that may provide support for the
possible outcomes is a {\em basic probability assignment (bpa)}, a function
that represents the impact of the piece of evidence on the subsets of
$\Theta$. A bpa assigns a number in the range [0,1] to each subset of
$\Theta$ such that the numbers sum to 1. The number assigned to the subset
$\Theta_1$ then denotes the amount of support the evidence directly
provides for the conclusions represented by $\Theta_1$.  When multiple
pieces of evidence are present, Dempster's combination rule is used to
compute a new bpa from the individual bpa's to represent their cumulative
effect.

The reasons for selecting the Dempster-Shafer theory as the basis for our
model are twofold. First, unlike the Bayesian model, it does not require a
complete set of {\em a priori} and conditional probabilities, which is
difficult to obtain for sparse pieces of evidence. Second, the
Dempster-Shafer theory distinguishes between situations in which no
evidence is available to support any conclusion and those in which equal
evidence is available to support each conclusion. Thus the outcome of the
model more accurately represents the {\em amount} of evidence available to
support a particular conclusion, i.e., the {\em provability} of the
conclusion \cite{pea_ur90}.

\subsection{Cues for Tracking Initiative}

In order to utilize the Dempster-Shafer theory for modeling initiative, we
must first identify the cues that provide evidence for initiative
shifts. Whittaker, Stenton, and Walker \cite{whi_ste_acl88,wal_whi_acl90}
have previously identified a set of utterance intentions that serve as cues
to indicate shifts or lack of shifts in initiative, such as prompts and
questions.  We analyzed our annotated TRAINS91 corpus and identified
additional cues that may have contributed to the shift or lack of shift in
task/dialogue initiatives during the interactions. This results in eight
cue types, which are grouped into three classes, based on the kind of
knowledge needed to recognize them. Table~\ref{cues} shows the three
classes, the eight cue types, their subtypes if any, whether a cue may
affect merely the dialogue initiative or both the task and dialogue
initiatives, and the agent expected to hold the initiative in the next
turn.

\begin{table*}[tb]
\footnotesize
\begin{center}
\begin{tabular}{|l|l|l|l|l|l|}
\hline
{\bf Class} & {\bf Cue Type} & {\bf Subtype} & {\bf Effect} & {\bf Initiative} 
& {\bf Example} \\ \hline
Explicit & Explicit requests & give up & both & hearer & {\em ``Any 
suggestions?'' ``Summarize the plan up to this point''} \\ \cline{3-6}
&                  & take over & both & speaker & {\em ``Let me handle this 
one.''} \\ \hline
Discourse & End silence & & both & hearer & \\ \cline{2-6}
& No new info & repetitions & both & hearer & {\em A: ``Grab 
the tanker, pick up oranges, go to Elmira,} \\
&                         & & & & \hspace{1.3em} {\em make them into orange
juice.''} \\
&                         & & & & {\em B: ``We go to Elmira, we make orange
juice, okay.''} \\ \cline{3-6}
&        & prompts & both & hearer & {\em ``Yeah'', ``Ok'', ``Right''}\\ 
\cline{2-6}
& Questions & domain & DI & speaker & {\em ``How far is it from Bath to
Corning?''} \\ \cline{3-6}
&           & evaluation & DI & hearer & {\em ``Can we do the route the 
banana guy isn't doing?''} \\ \cline{2-6}
& Obligation & task & both & hearer & {\em A: ``Any suggestions?''}
\\
& fulfilled         & & & & {\em B: ``Well, there's a boxcar at Dansville.''}
\\
&                     & & & & \hspace{1em} {\em ``But you have to change your
banana plan.''} \\
&                     & & & & {\em A: ``How long is it from Dansville to 
Corning?''} \\ \cline{3-6}
& & discourse & DI & hearer & {\em A: ``Go ahead and fill up E1 with 
bananas.''} \\
&&&&& {\em B: ``Well , we have to get a boxcar.''} \\
&&&&& {\em A: ``Right, okay. It's shorter to Bath from Avon.''} \\ \hline
Analytical & Invalidity & action & both & hearer &  {\em A: ``Let's get the
tanker car to Elmira and fill it with OJ.} \\
&           & & & & {\em B: ``You need to get oranges to the OJ factory.''} \\ \cline{3-6} 
&           & belief & DI & hearer &  {\em A: ``It's shorter to Bath from 
Avon.''} \\ 
&           & & & & {\em B: `` It's shorter to Dansville.''} \\
&           & & & & \hspace{1em} {\em ``The map is 
slightly misleading.''} \\ \cline{2-6}
& Suboptimality & & both & hearer & {\em A: ``Using Saudi on Thursday the 
eleventh.''} \\ 
&              & & & & {\em B: ``It's sold out.''} \\
&              & & & & {\em A: ``Is Friday open?''} \\
&              & & & & {\em B: ``Economy on Pan Am is open on Thursday.''} \\
\cline{2-6}
&Ambiguity & action & both & hearer & {\em A: ``Take one of the engines from 
Corning.''} \\
&          & & & & {\em B: ``Let's say engine E2.''} \\ \cline{3-6}
&          & belief & DI & hearer & {\em A: ``We would get back to Corning at 
4.''} \\
&          & & & & {\em B: ``4PM? 4AM?''} \\ \hline
\end{tabular}
\end{center}
\caption{Cues for Modeling Initiative}
\label{cues}
\vspace{1ex}
\hrule
\end{table*}

The first cue class, {\em explicit cues}, includes explicit requests by the
speaker to give up or take over the initiative. For instance, the utterance
``{\em Any suggestions?''} indicates the speaker's intention for the hearer
to take over both the task and dialogue initiatives. Such explicit cues can
be recognized by inferring the discourse and/or problem-solving intentions
conveyed by the speaker's utterances.

\looseness=-1000 The second cue class, {\em discourse cues}, includes cues
that can be recognized using linguistic and discourse information, such as
from the surface form of an utterance, or from the discourse relationship
between the current and prior utterances. It consists of four cue
types. The first type is perceptible silence at the end of an utterance,
which suggests that the speaker has nothing more to say and may intend to
give up her initiative. The second type includes utterances that do not
contribute information that has not been conveyed earlier in the
dialogue. It can be further classified into two groups: {\em repetitions},
a subset of the {\em informationally redundant utterances}
\cite{wal_coling92}, in which the speaker paraphrases an utterance by the
hearer or repeats the utterance verbatim, and {\em prompts}, in which the
speaker merely acknowledges the hearer's previous utterance(s). Repetitions
and prompts also suggest that the speaker has nothing more to say and
indicate that the hearer should take over the initiative
\cite{whi_ste_acl88}. The third type includes questions which, based on
anticipated responses, are divided into {\em domain} and {\em evaluation}
questions. {\em Domain} questions are questions in which the speaker
intends to obtain or verify a piece of domain knowledge.  They usually
merely require a direct response and thus typically do not result in an
initiative shift. {\em Evaluation} questions, on the other hand, are
questions in which the speaker intends to assess the quality of a proposed
plan. They often require an analysis of the proposal, and thus frequently
result in a shift in dialogue initiative. The final type includes
utterances that satisfy an outstanding task or discourse obligation. Such
obligations may have resulted from a prior request by the hearer, or from
an interruption initiated by the speaker himself. In either case, when the
task/dialogue obligation is fulfilled, the initiative may be reverted back
to the hearer who held the initiative prior to the request or interruption.

The third cue class, {\em analytical cues}, includes cues that cannot be
recognized without the hearer performing an evaluation on the speaker's
proposal using the hearer's private knowledge
\cite{chu_car_aaai94,chu_car_acl95}. After the evaluation, the hearer may
find the proposal {\em invalid}, {\em suboptimal}, or {\em ambiguous}. As a
result, he may initiate a subdialogue to resolve the problem, resulting in
a shift in task/dialogue initiatives.\footnote{Whittaker, Stenton, and
Walker treat subdialogues initiated as a result of these cues as
interruptions, motivated by their collaborative planning principles
\cite{whi_ste_acl88,wal_whi_acl90}.}

\subsection{Utilizing the Dempster-Shafer Theory}

\looseness=-1000 As discussed earlier, at the end of each turn, new
task/dialogue initiative indices are computed based on the current indices
and the effect of the observed cues to determine the next task/dialogue
initiative holders. In terms of the Dempster-Shafer theory, new
task/dialogue bpa's ($m_{t-new}$/$m_{d-new}$)\footnote{Bpa's are
represented by functions whose names take the form of $m_{sub}$. The
subscript {\em sub} may be {\em t-X} or {\em d-X}, indicating that the
function represents the task or dialogue bpa under scenario X.} are
computed by applying Dempster's combination rule to the bpa's representing
the current initiative indices\footnote{The initiative indices are
represented as bpa's. For instance, the current task initiative indices
take the following form: $m_{t-cur}(speaker) = x$ and $m_{t-cur}(hearer) =
1-x$.} and the bpa of each observed cue.

Evidently, some cues provide stronger evidence for an initiative shift than
others. Furthermore, a cue may provide stronger support for a shift in
dialogue initiative than in task initiative. Thus, we associate with each
cue two bpa's to represent its effect on changing the current task and
dialogue initiative indices, respectively. We extended our annotations of
the TRAINS91 dialogues to include, in addition to the agent(s) holding the
task and dialogue initiatives for each turn, a list of cues observed during
that turn. Initially, each cue$_i$ is assigned the following bpa's:
$m_{t-i}(\Theta) = 1$ and $m_{d-i}(\Theta) = 1$, where $\Theta$ =
\{speaker,hearer\}. In other words, we assume that the cue has no effect on
changing the current initiative indices. We then developed a training
algorithm ({\bf Train-bpa}, Figure~\ref{train_algorithm}) and applied it on
the annotated data to obtain the final bpa's.

\begin{figure}[htb]
\footnotesize

\noindent {\bf Train-bpa}(annotated-data):

\begin{algorithm}

\item $m_{t-cur} \leftarrow$ default task initiative indices

      $m_{d-cur} \leftarrow$ default dialogue initiative indices

      cur-data $\leftarrow$ {\bf read}(annotated-data)

      cue-set $\leftarrow$ cues in cur-data 

\item \label{compute} {\em /* compute new initiative indices */}

      $m_{t-obs} \leftarrow$ task initiative bpa's for cues in
      cue-set

      $m_{d-obs} \leftarrow$ dialogue initiative bpa's for cues in
      cue-set

      $m_{t-new} \leftarrow$ {\bf combine}($m_{t-cur},m_{t-obs}$)
  
      $m_{d-new} \leftarrow$ {\bf combine}($m_{d-cur},m_{d-obs}$)

\item \label{predict} {\em /* determine predicted next initiative holders */}

      If $m_{t-new}(speaker) \ge m_{t-new}(hearer)$, 

      \hspace*{.5em} t-predicted $\leftarrow$ speaker

      Else, t-predicted $\leftarrow$ hearer

      If $m_{d-new}(speaker) \ge m_{d-new}(hearer)$, 

      \hspace*{.5em} d-predicted $\leftarrow$ speaker

      Else, d-predicted $\leftarrow$ hearer

\item \label{adjust} {\em /* find actual initiative holders and compare */}

      new-data $\leftarrow$ {\bf read}(annotated-data)

      t-actual $\leftarrow$ actual task initiative holder in new-data
 
      d-actual $\leftarrow$ actual dialogue initiative holder in new-data

      If t-predicted $\neq$ t-actual, 

      \hspace*{.5em} {\bf Adjust-bpa}(cue-set,task)

      \hspace*{.5em} {\bf Reset-current-bpa}($m_{t-cur}$)

      If d-predicted $\neq$ d-actual,

      \hspace*{.5em} {\bf Adjust-bpa}(cue-set,dialogue)

      \hspace*{.5em} {\bf Reset-current-bpa}($m_{d-cur}$)

\item If end-of-dialogue, return
      
      Else, \hspace*{1em} {\em /* swap roles of speaker and hearer */}

      \hspace*{.5em} $m_{t-cur}(speaker) \leftarrow m_{t-new}(hearer)$

      \hspace*{.5em} $m_{d-cur}(speaker) \leftarrow m_{d-new}(hearer)$

      \hspace*{.5em} $m_{t-cur}(hearer) \leftarrow m_{t-new}(speaker)$

      \hspace*{.5em} $m_{d-cur}(hearer) \leftarrow m_{d-new}(speaker)$ 

      \hspace*{.5em} cue-set $\leftarrow$ cues in new-data

      \hspace*{.5em} Goto step~\ref{compute}.

\end{algorithm}
\caption{Training Algorithm for Determining BPA's}
\label{train_algorithm}
\vspace{1ex}
\hrule
\end{figure}

For each turn, the task and dialogue bpa's for each observed cue are used,
along with the current initiative indices, to determine the new initiative
indices (step~\ref{compute}). The {\bf combine} function utilizes
Dempster's combination rule to combine pairs of bpa's until a final bpa is
obtained to represent the cumulative effect of the given bpa's. The
resulting bpa's are then used to predict the task/dialogue initiative
holders for the next turn (step~\ref{predict}). If this prediction
disagrees with the actual value in the annotated data, {\bf Adjust-bpa} is
invoked to alter the bpa's for the observed cues, and {\bf
Reset-current-bpa} is invoked to adjust the current bpa's to reflect the
actual initiative holder (step~\ref{adjust}).

\begin{figure*}[htb]
\begin{center}
\hfil
\subfigure[Task Initiative
Prediction]{\psfig{figure=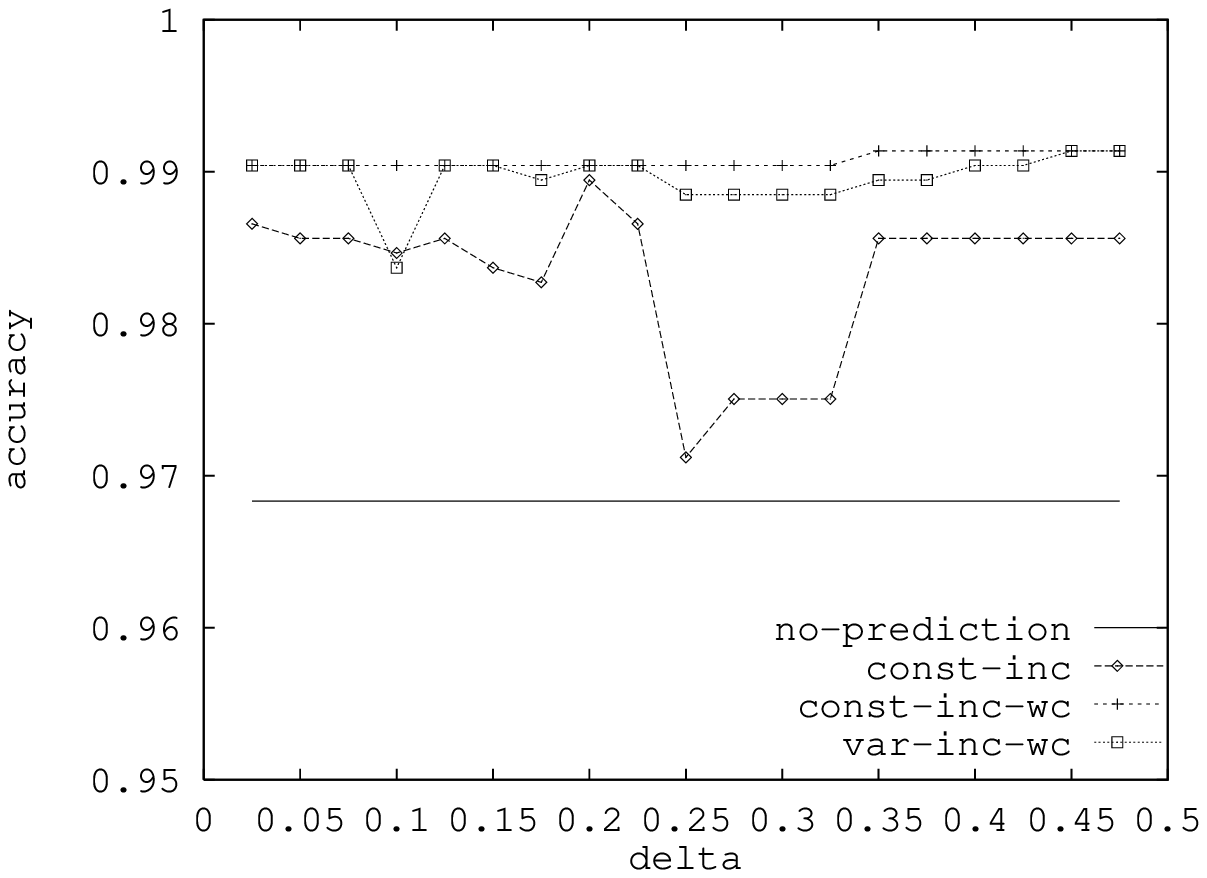,width=2.7in}}
\hfil
\subfigure[Dialogue Initiative 
Prediction]{\psfig{figure=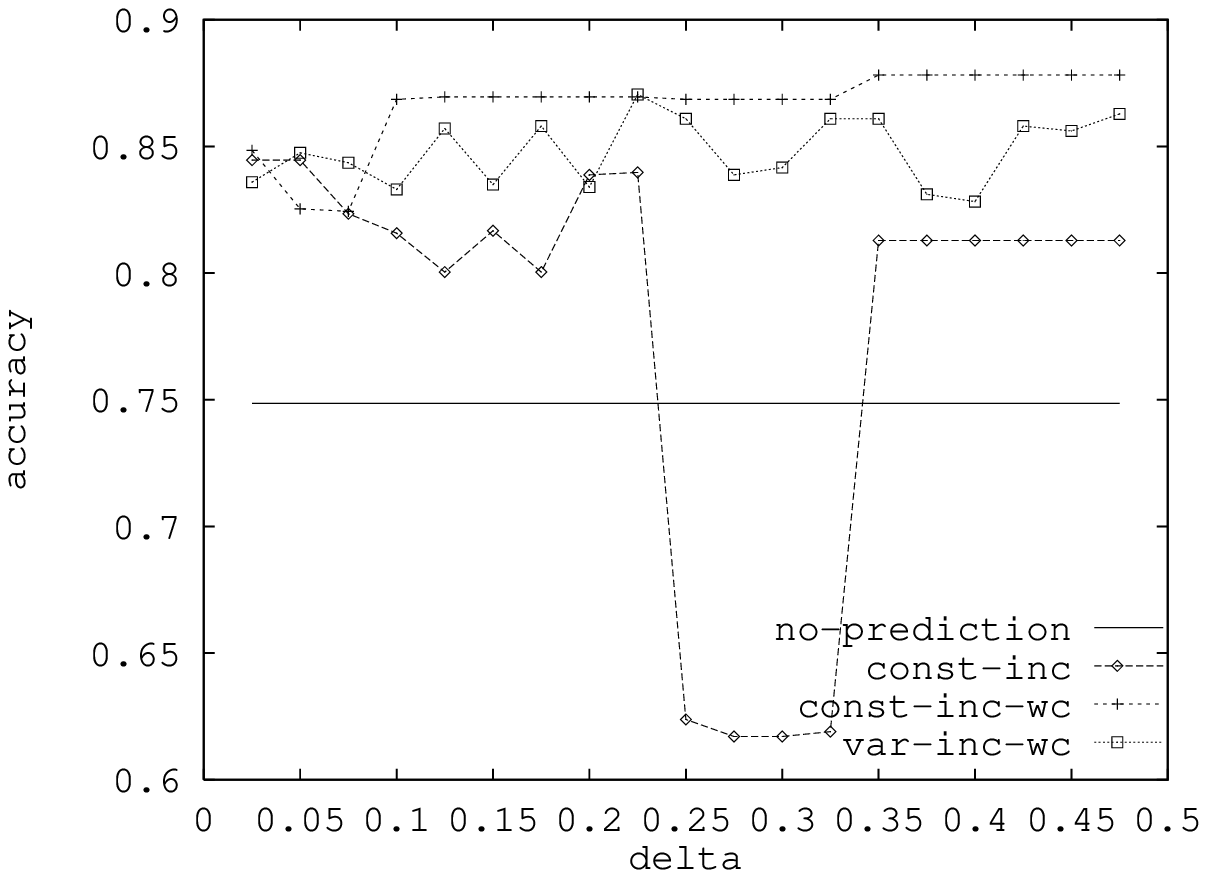,width=2.7in}}
\hfil
\end{center}
\caption{Comparison of Three Adjustment Methods}
\label{plots}
\vspace{1ex}
\hrule
\end{figure*}

{\bf Adjust-bpa} adjusts the bpa's for the observed cues in favor of the
actual initiative holder. We developed three adjustment methods by varying
the effect that a disagreement between the actual and predicted initiative
holders will have on changing the bpa's for the observed cues.  The first
is {\em constant-increment} where each time a disagreement occurs, the
value for the actual initiative holder in the bpa is incremented by a
constant ($\Delta$), while that for $\Theta$ is decremented by
$\Delta$. The second method, {\em constant-increment-with-counter},
associates with each bpa for each cue a counter which is incremented when a
correct prediction is made, and decremented when an incorrect prediction is
made. If the counter is negative, the {\em constant-increment} method is
invoked, and the counter is reset to 0. This method ensures that a bpa will
only be adjusted if it has no ``credit'' for correct predictions in the
past. The third method, {\em variable-increment-with-counter}, is a
variation of {\em constant-increment-with-counter}. However, instead of
determining whether an adjustment is needed, the counter determines the
amount to be adjusted. Each time the system makes an incorrect prediction,
the value for the actual initiative holder is incremented by
$\Delta/2^{count+1}$, and that for $\Theta$ decremented by the same amount.

In addition to experimenting with different adjustment methods, we also
varied the increment constant, $\Delta$. For each adjustment method, we ran
19 training sessions with $\Delta$ ranging from 0.025 to 0.475,
incrementing by 0.025 between each session, and evaluated the system based
on its accuracy in predicting the initiative holders for each turn. We
divided the TRAINS91 corpus into eight sets based on speaker/hearer
pairs. For each $\Delta$, we cross-validated the results by applying the
training algorithm to seven dialogue sets and testing the resulting bpa's
on the remaining set.  Figures~\ref{plots}(a) and \ref{plots}(b) show our
system's performance in predicting the task and dialogue initiative
holders, respectively, using the three adjustment methods.\footnote{For
comparison purposes, the straight lines show the system's performance
without the use of cues, i.e., always predict that the initiative remains
with the current holder.}

\subsection{Discussion}

Figure~\ref{plots} shows that in the vast majority of cases, our prediction
methods yield better results than making predictions without cues.
Furthermore, substantial improvement is gained by the use of counters since
they prevent the effect of the ``exceptions of the rules'' from
accumulating and resulting in erroneous predictions. By restricting the
increment to be inversely exponentially related to the ``credit'' the bpa
had in making correct predictions, {\em variable-increment-with-counter}
obtains better and more consistent results than {\em
constant-increment}. However, the exceptions of the rules still resulted in
undesirable effects, thus the further improved performance by {\em
constant-increment-with-counter}.

\begin{table*}[tb]
\footnotesize
\hfil
\begin{center}
\subfigure[Task Initiative Errors]{
\begin{tabular}{|l|l|c|c|c|c|}
\hline
{\bf Cue Type} & {\bf Subtype} & \multicolumn{2}{|c|}{\bf Shift} 
& \multicolumn{2}{|c|}{\bf No-Shift} \\ \cline{3-6}
& & error & total & error & total \\ \hline
Invalidity & action & 2 & 3 & 0 & 11 \\ \hline
Suboptimality & & 1 & 1 & 0 & 0 \\ \hline
Ambiguity & action & 3 & 7 & 1 & 5 \\ \hline
\end{tabular}}
\end{center}
\hfil
\begin{center}
\subfigure[Dialogue Initiative Errors]{
\begin{tabular}{|l|l|c|c|c|c|}
\hline
{\bf Cue Type} & {\bf Subtype} & \multicolumn{2}{|c|}{\bf Shift} 
& \multicolumn{2}{|c|}{\bf No-Shift} \\ \cline{3-6}
& & error & total & error & total \\ \hline
End silence & & 13 & 41 & 0 & 53 \\ \hline
No new info & prompts & 7 & 193 & 1 & 6 \\ \hline
Questions & domain & 13 & 31 & 0 & 98 \\ \cline{2-6}
          & evaluation & 8 & 28 & 5 & 7 \\ \hline
Obligation fulfilled & discourse & 12 & 198 & 1 & 5 \\ \hline
Invalidity & & 11 & 34 & 0 & 0 \\ \hline
Suboptimality & & 1 & 1 & 0 & 0 \\ \hline
Ambiguity & & 9 & 24 & 0 & 0 \\ \hline
\end{tabular}}
\end{center}
\hfil
\caption{Summary of Prediction Errors}
\label{errors}
\vspace{1ex}
\hrule
\end{table*}

We analyzed the cases in which the system, using {\em
constant-increment-with-counter} with $\Delta$ = .35,\footnote{This is the
value that yields the optimal results (Figure~\ref{plots}).} made
erroneous predictions. Tables~\ref{errors}(a) and \ref{errors}(b) summarize
the results of our analysis with respect to task and dialogue initiatives,
respectively. For each cue type, we grouped the errors based on whether or
not a shift occurred in the actual dialogue. For instance, the first row
in Table~\ref{errors}(a) shows that when the cue {\em invalid action} is
detected, the system failed to predict a task initiative shift in 2 out of
3 cases. On the other hand, it correctly predicted all 11 cases where no
shift in task initiative occurred.  Table~\ref{errors}(a) also shows that
when an analytical cue is detected, the system correctly predicted all but
one case in which there was no shift in task initiative. However, 55\% of
the time, the system failed to predict a shift in task
initiative.\footnote{In the case of suboptimal actions, we encounter the
sparse data problem. Since there is only one instance of the cue in the set
of dialogues, when the cue is present in the testing set, it is absent from
the training set.}  This suggests that other features need to be taken into
account when evaluating user proposals in order to more accurately model
initiative shifts resulting from such cues. Similar observations can be
made about the errors in predicting dialogue initiative shifts when
analytical cues are observed (Table~\ref{errors}(b)).

\looseness=-1000
Table~\ref{errors}(b) shows that when a perceptible silence is detected at
the end of an utterance, when the speaker utters a prompt, or when an
outstanding discourse obligation is fulfilled (first three rows in table),
the system correctly predicted the dialogue initiative holder in the vast
majority of cases. However, for the cue class {\em questions}, when the
actual initiative shift differs from the norm, i.e., speaker retaining
initiative for evaluation questions and hearer taking over initiative for
domain questions, the system's performance worsens. In the case of domain
questions, errors occur when 1) the response requires more reasoning than
do typical domain questions, causing the hearer to take over the dialogue
initiative, or 2) the hearer, instead of merely responding to the question,
offers additional helpful information. In the case of evaluation questions,
errors occur when 1) the result of the evaluation is readily available to
the hearer, thus eliminating the need for an initiative shift, or 2) the
hearer provides extra information. We believe that although it is difficult
to predict when an agent may include extra information in response to a
question, taking into account the cognitive load that a question places on
the hearer may allow us to more accurately predict dialogue initiative
shifts.

\section{Applications in Other Environments}
\label{apps}

To investigate the generality of our system, we applied our training
algorithm, using the {\em constant-increment-with-counter} adjustment
method with $\Delta$ = 0.35, on the TRAINS91 corpus to obtain a set of
bpa's. We then evaluated the system on subsets of dialogues from four other
corpora: the TRAINS93 dialogues \cite{trains93a}, airline reservation
dialogues \cite{sri92}, instruction-giving dialogues \cite{maptask96}, and
non-task-oriented dialogues \cite{switchboard_cc93}. In addition, we
applied our baseline strategy which makes predictions without the use of
cues to each corpus.

\begin{table*}[tb]
\footnotesize
\begin{center}
\begin{tabular}{|l|c|c|c|c|c|c|c|c|c|c|}
\hline
{Corpus} & \multicolumn{2}{|c|}{TRAINS91 (1042)}
& \multicolumn{2}{|c|}{TRAINS93 (256)} &
\multicolumn{2}{|c|}{Airline (332)} & \multicolumn{2}{|c|}{Maptask (320)} &
\multicolumn{2}{|c|}{Switchboard (282)} \\ \cline{2-11}
(\# turns) & task & dialogue & task & dialogue & task & dialogue & task 
& dialogue & task & dialogue \\ \hline
Expert & 41 & 311 & 37 & 101 & 194 & 193 & 320 & 277 & N/A & 166 \\
control & (3.9\%) & (29.8\%) & (14.4\%) & (39.5\%) & (58.4\%) & (58.1\%) 
& (100\%) & (86.6\%) & & (59.9\%) \\ \hline
No cue & 1009 & 780 & 239 & 189 & 308 & 247 & 320 & 270 & N/A
& 193  \\ 
& (96.8\%) & (74.9\%) & (93.3\%) & (73.8\%) & (92.8\%) & (74.4\%) & (100\%) 
& (84.4\%) & & (68.4\%) \\ \hline
{\em const-inc-} & 1033 & 915 & 250 & 217 & 316 & 281 & 320 & 297
& N/A & 216  \\ 
{\em w-count} & (99.1\%) & (87.8\%) & (97.7\%) & (84.8\%) & (95.2\%) 
& (84.6\%) & (100\%) & (92.8\%) && (76.6\%)\\ \hline
{\em Improvement} & 2.3\% & 12.9\% & 4.4\% & 11.0\% & 2.4\% & 10.2\% & 0.0\% 
& 8.4\% & N/A & 8.2\% \\ \hline
\end{tabular}
\end{center}
\caption{Comparison Across Different Application Environments}
\label{domains}
\vspace{1ex}
\hrule
\end{table*}

\looseness=-1000 Table~\ref{domains} shows a comparison between the
dialogues from the five corpora and the results of this evaluation. Row 1
in the table shows the number of turns where the {\em expert}\footnote{The
{\em expert} is assigned as follows: in the TRAINS domain, the system; in
the airline domain, the travel agent; in the maptask domain, the
instruction giver; and in the switchboard dialogues, the agent who holds
the dialogue initiative the majority of the time.} holds the task/dialogue
initiative, with percentages shown in parentheses. This analysis shows that
the distribution of initiatives varies quite significantly across corpora,
with the distribution biased toward one agent in the TRAINS and maptask
corpora, and split fairly evenly in the airline and switchboard
dialogues. Row 2 shows the results of applying our baseline prediction
method to the various corpora. The numbers shown are correct predictions in
each instance, with the corresponding percentages shown in
parentheses. These results indicate the difficulty of the prediction
problem in each corpus that the task/dialogue initiative distribution (row
1) fails to convey. For instance, although the dialogue initiative is
distributed approximately 30/70\% between the two agents in the TRAINS91
corpus and 40/60\% in the airline dialogues, the prediction rates in row 2
shows that in both cases, the distribution is the result of shifts in
dialogue initiative in approximately 25\% of the dialogue turns. 
Row 3 in the table shows the prediction results when applying our
training algorithm using the {\em constant-increment-with-counter}
method. Finally, the last row shows the improvement in percentage points
between our prediction method and the baseline prediction method.
To test the statistical significance of the differences between the
results obtained by the two prediction algorithms, for each corpus, we
applied Cochran's $Q$ test \cite{coc_bio50} to the results in rows 2 and
3. The tests show that for all corpora, the differences between the two
algorithms when predicting the task and dialogue initiative holders are
statistically significant at the levels of p$<$0.05 and p$<10^{-5}$,
respectively.

Based on the results of our evaluation, we make the following
observations. First, Table~\ref{domains} illustrates the generality of our
prediction mechanism. Although the system's performance varies across
environments, the use of cues consistently improves the system's accuracies
in predicting the task and dialogue initiative holders by 2-4 percentage
points (with the exception of the maptask corpus in which there is no room
for improvement)\footnote{In the maptask domain, the task initiative
remains with one agent, the instruction giver, throughout the dialogue.}
and 8-13 percentage points, respectively. Second, Table~\ref{domains} shows
the specificity of the trained bpa's with respect to application
environments. Using our prediction mechanism, the system's performances on
the collaborative planning dialogues (TRAINS91, TRAINS93, and airline
reservation) most closely resemble one another (last row in table). This
suggests that the bpa's may be somewhat sensitive to application
environments since they may affect how agents interpret cues. Third, our
prediction mechanism yields better results on task-oriented dialogues. This
is because such dialogues are constrained by the goals; therefore, there
are fewer digressions and offers of unsolicited opinion as compared to the
switchboard corpus.

\section{Conclusions}

This paper discussed a model for tracking initiative between participants
in mixed-initiative dialogue interactions. We showed that distinguishing
between task and dialogue initiatives allows us to model phenomena in
collaborative dialogues that existing systems are unable to explain. We
presented eight types of cues that affect initiative shifts in dialogues,
and showed how our model predicts initiative shifts based on the current
initiative holders and and the effects that observed cues have on changing
them. Our experiments show that by utilizing the {\em
constant-increment-with-counter} adjustment method in determining the basic
probability assignments for each cue, the system can correctly predict the
task and dialogue initiative holders 99.1\% and 87.8\% of the time,
respectively, in the TRAINS91 corpus, compared to 96.8\% and 74.9\% without
the use of cues. The differences between these results are shown to be
statistically significant using Cochran's Q test. In addition, we
demonstrated the generality of our model by applying it to dialogues in
different application environments. The results indicate that although the
basic probability assignments may be sensitive to application environments,
the use of cues in the prediction process significantly improves the
system's performance.

\section*{Acknowledgments}

We would like to thank Lyn Walker, Diane Litman, Bob Carpenter, and
Christer Samuelsson for their comments on earlier drafts of this paper, Bob
Carpenter and Christer Samuelsson for participating in the coding
reliability test, as well as Jan van Santen and Lyn Walker for discussions
on statistical testing methods.

\end{document}